%% file: main.tex
\documentclass{article}

\usepackage{microtype}
\usepackage{graphicx}
\usepackage{booktabs} %

\usepackage{url}
\usepackage{verbatim}
\usepackage{enumitem}

\usepackage{listings}
\usepackage{amsmath}
\usepackage{amsfonts}
\usepackage{subcaption}
\usepackage{multirow}
\usepackage[font=small]{caption}
\usepackage{algorithmic}
\usepackage[linesnumbered,ruled,vlined]{algorithm2e}
\usepackage{float}
\usepackage{pgfplots}
\pgfplotsset{compat=1.15}
\usepackage{filecontents}
\usepackage{keyval}
\usepackage{tikz}
\usetikzlibrary{shapes.geometric,positioning}

\newcommand{\std}{\scriptsize $\pm$}

\usepackage{hyperref}

\newcommand{\cutspaceabove}{\vspace{-0.5em}}
\newcommand{\cutspacebelow}{\vspace{-0.4em}}
\newcommand{\cuttablebelow}{\vspace{-0.3em}}
\newcommand{\cutspacepara}{\vspace{-0.8em}}

\usepackage[final]{neurips_2020}

\title{Few-shot Sequence Learning with Transformers}

\author{
  Lajanugen Logeswaran$^1$,
  Ann Lee$^2$,
  Myle Ott$^2$,
  Honglak Lee$^{1}$,
  \\
  \textbf{Marc'Aurelio Ranzato}$^2$\textbf{,}
  \textbf{Arthur Szlam}$^2$ 
  \\
  $^1$University of Michigan,
  $^2$Facebook AI Research
}

\begin{document}

\maketitle

\input{sections/abstract}

\input{sections/intro}
\input{sections/problem}

\input{sections/approach}

\input{sections/relatedv2}

\input{sections/experiments}

\input{sections/conclusion}

\bibliography{main}
\bibliographystyle{icml2019}

\newpage
\appendix

\input{sections/supplement}

\end{document}

%% file: sections/abstract.tex
\begin{abstract}
Few-shot algorithms aim at learning new tasks provided only a handful of training examples.
In this work we investigate few-shot learning in the setting where the data points are sequences of tokens and propose an efficient learning algorithm based on Transformers.
In the simplest setting, we append a token to an input sequence which represents the particular task to be undertaken, 
and show that the embedding of this token can be optimized on the fly given few labeled examples.
Our approach does not require complicated changes to the model architecture such as adapter layers nor computing second order derivatives as is currently popular in the meta-learning and few-shot learning literature.
We demonstrate our approach on a variety of tasks, and analyze the generalization properties of several model variants and baseline approaches.  
In particular, we show that compositional task descriptors can improve performance.
Experiments show that our approach works at least as well as other methods, while being more computationally efficient.
\end{abstract}

%% file: sections/intro.tex
\section{Introduction}

The problem of learning a classifier from a handful examples has received considerable attention in the vision domain under the name of {\em few-shot learning}~\citep{fink05, 1shot}.
However, less work exists in the space of few-shot problems involving {\em discrete} sequences, such as sequences of discrete actions in reinforcement learning or
sequences of words in natural language processing. %
In this work, we study the problem of sequence classification and modeling in the few-shot regime. 
Specifically, we assume there are several training tasks available for learning and, at test time, we are interested in performing few-shot adaptation to a given new task.

Transformers~\citep{vaswani2017attention} have been very successful at modeling discrete sequences~\citep{barrault-etal-2019-findings, devlin2018bert, parisotto19}.  
Further, they have been shown to use context tokens appended to an input to adapt their generations or to switch between different tasks \citep{lample19rewrite,shen19,GROVER, keskar2019ctrl}.  
Thus, one might hope that such context tokens could be effectively used in the meta-learning setting for discrete sequences.

In this work, we show that this is indeed the case.
Our approach to few-shot learning introduces a set of task specific parameters (a \textit{task embedding}), in addition to the parameters of the model that are shared among all tasks.
Unlike other approaches that require architectural changes~\citep{houlsby2019parameter}, task embeddings are simply fed to the input of the transformer.
Learning a new task consists of inferring an appropriate task embedding for the task, leaving the shared model parameters intact. 
Towards this end, we propose a simple training algorithm where the task embedding is found via gradient based optimization, which is simpler and computationally less expensive than second order 
optimization methods~\citep{finn2017model, zintgraf2018caml}.

To summarize, our contributions in this work are as follows. 
First, we show that a simple alternating-minimization approach for few-shot learning works well in combination with the transformer architecture. 
Second, we show that a simple yet effective way to condition the transformer with task information is via input conditioning (i.e., feeding task information as input to the transformer); this naturally extends to compositional task information. 
Third, we introduce a battery of synthetic sequence classification and modeling tasks to benchmark in a controlled setting various baseline approaches 
and model variants for few shot learning of discrete sequences.
And finally, we demonstrate that the proposed approach offers a better trade-off between few-shot performance 
and run time cost compared to other baselines, including meta-learning approaches.

%% file: sections/problem.tex
\cutspaceabove
\section{Problem Definition}
\cutspacebelow
We assume a distribution $p_\text{data}(\mathcal{T})$ over tasks from which disjoint sets of training, validation and test sets of tasks are drawn.
The set of training tasks is denoted by $\{\mathcal{T}_i^\text{train}\}_{i=1}^{N}$, where each task $\mathcal{T}_i^\text{train}$ has an associated set of training examples $\{ (x^i_j, y^i_j)_{j=1}^{N_i} \}$.
Validation and test tasks are defined similarly, except each test task only has $k$ training examples.
We use the training tasks to learn the model parameters and the validation tasks to determine hyperparameters.
The optimal parameters and hyperparameters identified are used for evaluating average model performance on test tasks.
This involves first (optionally) training on the small set of training examples accompanied by a test task, followed by testing on the corresponding test set.

We focus on two types of tasks that involve discrete sequences as inputs and outputs: sequence classification and transduction.
In {\em sequence classification}, the inputs $x$ are sequences and the output $y$ is a discrete categorical label.
In {\em sequence transduction}, each task consists of modeling the joint distribution of a sequence $y$, conditioned on some input context sequence $x$.
The performance metrics for these settings are respectively accuracy and perplexity, averaged across the test tasks.

%% file: sections/approach.tex
\cutspaceabove
\section{Approach}
\cutspacebelow

\subsection{Architecture}

In this work, we explore an adaptation of transformers to  the few-shot regime.  
Previous works have shown that the behavior of a transformer model can be conditioned by 
appending special tokens describing the task to be performed on the input sequence~\citep{lample19rewrite,GROVER}. 
We append a \textit{task embedding} vector which represents information about the task of interest to the input sequence of token embeddings.
We intend to control the overall behavior of the model for a particular task by altering the task embeddings while keeping the rest of the model parameters intact.

For classification tasks, we use a transformer encoder similar to the BERT model~\citep{devlin2018bert}.
A classification head sits on the final layer representation of a special token at the beginning of the sequence. 
We replace this special token with the task embedding vector $z$ in our model.
We use a transformer decoder architecture for the transduction tasks and append the task embedding vector to the input sequence similar to the classification setting.

In both settings we compute a log-likelihood of the form $\log p(y | x, z; \theta)$,
where $x$ is the input sequence, $z$ is the task embedding and $\theta$ the model parameters. 
In the classification setting $y$ is a single categorical value.
In the sequence transduction setting, $y$ is a sequence and the log likelihood decomposes 
as the sum of the conditional log-likelihood terms via the chain rule of probability theory: 
$\log p(y | x, z; \theta) = \sum_i \log p(y_i | y_{i-1}, \cdots, y_1, x, z; \theta)$.

Note that in practical applications $\theta$ can be high-dimensional, in the order of hundreds of millions. 
Our goal is to alleviate overfitting in the few-shot regime by adapting only $z$ to learn a new task, where $z$ is a small vector with at most a few hundred components.
Next, we describe how we learn the model parameters $\theta$ and how we estimate the task embedding $z$ for a given task.

\input{figs/algorithm}

\subsection{Training and Inference Algorithm} \label{training}
We train our models with an alternating-minimization scheme similar to~\citet{maurer2013sparse} and~\citet{kumar2012learning}, 
that can be considered a simplification of the CAVIA approach in~\citet{zintgraf2018caml} (or as a refinement of the ``first-order'' method in that work). 
See Algorithm~\ref{alg:training} for pseudo-code.
We separate the weights of the network defining the model into shared weights $\theta$, and per-task weights, as in CAVIA.
In our case, the per-task weights form the embedding $z$, one for each task; while all other parameters $\theta$ are shared. 

Given a few examples from the training task $\mathcal{T}_i^\text{train}$ (line 3), we alternate training $z_{\mathcal{T}_i^\text{train}}$ (the task embedding of the $\mathcal{T}_i^\text{train}$ task) for a few gradient descent steps keeping $\theta$ fixed (see line 5 and 6), and then update $\theta$ based on the optimal task embedding.
In practice, however, we found it helpful to update $\theta$ based on gradients accumulated for the intermediate values of the task embedding encountered in the inner loop optimization (line 7).
We surmise that this optimization choice helps the model finding better task embeddings as the whole parameter vector $\theta$ is updated to account for this search.
Task embedding gradient updates (line 6) are performed until the loss no longer improves or the maximum number of update steps has been reached.
Note that unlike prior methods such as MAML or CAVIA we do not backpropagate gradients through an optimization process, which simplifies and speeds up our optimization.
We call our method, transformer trained with Alternating Minimization (TAM) -- although the alternating minimization algorithm could be applied to other
architectures as well.
At test time, given a new task $\mathcal{T}^\text{test}$, $z_{\mathcal{T}^\text{test}}$ is trained with a few steps of gradient descent (similar to line 6), with all other parameters held fixed. 
Since TAM is trained to optimize task embeddings on the fly, we expect it to find good embeddings of the new task 
at test time as well.

%% file: figs/algorithm.tex
\begin{algorithm}[!t]
    \setcounter{AlgoLine}{0}
    \SetKwInOut{Input}{Input}
    \SetKwInOut{Output}{Output}

    \Input{Training tasks $\mathcal{T}_1^\text{train}, ..., \mathcal{T}_N^\text{train}$}
    \Output{Model parameters $\theta$}

    \Repeat{max training iterations}{
      Sample a training task: $\mathcal{T}_i^\text{train}$
      
      Sample $N_i \ge k$ training examples from the task 
      $\{(x^j,y^j)_{j=1,\cdots,N_i}\} \sim \mathcal{T}_i^\text{train}$
      
      \vspace{0.4em} Initialize: $z_{\mathcal{T}_i^\text{train}} = 0, \Delta \theta = 0$
      
      \While{loss improves and max number of updates not reached }
      {
 $z_{\mathcal{T}_i^\text{train}} {\scriptsize \leftarrow} z_{\mathcal{T}_i^\text{train}} - \nabla_{z_{\mathcal{T}_i^\text{train}}} \sum_j  \, - \log p(y^j | x^j, z_{\mathcal{T}_i^\text{train}}; \theta)$ 
        
        $\Delta \theta \leftarrow \Delta \theta - \nabla_\theta \sum_{j=1}^{N_i}  \, - \log p(y^j | x^j, z_{\mathcal{T}_i^\text{train}}; \theta)$
      }
      $\theta  \leftarrow \theta + \Delta \theta $
    }
    \caption{\textbf{TAM for k-shot Learning}}
    \label{alg:training}
    \cuttablebelow
\end{algorithm}

%% file: sections/relatedv2.tex
\cutspaceabove
\section{Related work}
\cutspacebelow

\paragraph{Few-shot learning and meta-learning} 
There is now a vast literature on learning methods designed for quickly adapting to new settings.   
At a  coarse level, one can consider classes of methods that adapt the learning {\it algorithm} 
based on the task (and so are ``meta-learners'')~\citep{schmidhuber1987evolutionary,hochreiter2001learning, andrychowicz2016learning,finn2017model,nichol2018reptile}, or describe {\it model architectures} that can adapt to learn sample-efficiently over a task distribution~\citep{vinyals2016matching, snell2017prototypical}.   
Many methods have elements of both of these, e.g. \cite{mishra2017simple, rusu2018meta, zintgraf2018caml}.  

The method we describe in this work can be considered squarely in the class of model architectures for sample efficient learning. 
It is a descendant of \citet{hinton1987using, schmidhuber1992learning, ba2016using} and is closely related 
to~\citet{rusu2018meta, zintgraf2018caml} in that we pick a subset of the weights of the model that are task 
specific (the ``fast'' weights), and update them using the training examples for a specific task; 
but update the other (``slow'') weights on all training examples for all tasks.   
Our approach is closest to~\citet{zintgraf2018caml}, but differs in the way the fast weights are used by the model. 
We do not use higher-order gradients for the slow weights, instead we use an alternating minimization type update.

\cutspaceabove
\paragraph{Task transfer for transformers}  Our approach is also related to other recent work in natural language processing.  
We leverage the particular structure of the transformer architecture~\citep{vaswani2017attention}, 
which has been successful in many NLP tasks.
Several works have shown that adding a token to an input can be used to switch between different 
tasks \citep{lample19rewrite,shen19,GROVER, keskar2019ctrl}. %
Transformer language models trained on large corpora have also been recently shown to have impressive few-shot learning capabilities \citep{brown2020language}.
More generally, with the success of methods based on pretraining transformer models~\citep{devlin2018bert}, and finetuning on target tasks, there have been  several works discussing how to adapt a pre-trained model without full finetuning~\citep{houlsby2019parameter, stickland2019bert} but their focus has been on reducing the number of parameters subject to optimization at finetuning time as opposed to reducing the number of examples as in this study. %

%% file: sections/experiments.tex
\section{Experiments}
\cutspacebelow

\subsection{Model and Training Details}
\cutspacebelow
In the classification setting, TAM is a bidirectional transformer that takes the input sequence $x$ and the task embedding $z$ as input, and outputs a distribution over classes. 
In the sequence transduction setting, TAM is a transformer decoder with a causal attention mechanism and takes as additional input the output sequence $y$ up to the token before the last.
In this case the model is trained to predicted the sequence $y$ at the last $|y|$ (length of sequence $y$) time steps.
Since model parameters are shared across tasks, TAM needs to leverage the task embedding to perform the tasks well. 
Both classification and transduction models are trained with cross-entropy loss.

Unless otherwise specified our transformer has 4 layers with an embedding size of 128.
We use the Adam optimizer~\citep{kingma2014adam} for both outer and inner loop optimization.
The maximum number of task embedding optimization steps is set to 25 during training.
We train a single model using $N$ samples at training time, treating $N$ as a hyperparameter and apply it to k-shot problems with different values of $k$ at test time.
The size of the task embedding was set to match the embedding dimension of the transformer (128).
We discuss more details about hyperparameter choices and how they influence model performance in section \ref{sec:discussion}.

\subsection{Baselines}
\cutspacebelow

\paragraph{Task-Agnostic transformer} 
This baseline uses the same architecture as TAM but is not informed about the existence of different tasks at training time, i.e., no task embedding is fed at the input. 
At test time, the model is fine-tuned on the $k$ training examples from the test task.
\cutspacepara

\paragraph{Multitask transformer} 
This is a transformer that is conditioned on the current task both in the classification and transduction settings. 
It is identical to TAM except all parameters including task embeddings are trained by standard back-propagation, without any alternating minimization.
\cutspacepara

\paragraph{Matching Networks~\citep{vinyals2016matching}} 
We consider Matching Networks only in our classification setting, as it is not straightforward to use it for transduction.
We use a transformer to model the similarity between a query instance and support set instance which takes the concatenation of the two sequences as input and outputs a similarity score.
The prediction is a convex sum of the training example labels, the weights being similarity scores.
\cutspacepara

\paragraph{SNAIL~\citep{mishra2017simple}} \hspace{-0.7em}
This model is similar to the task-agnostic transformer except the input is augmented with the concatenation of all input-output training pairs. 
For both Matching Networks and SNAIL, we construct training episodes by sampling $k$ training examples to define a task, to match the test scenario.
We train different models for each $k$-shot problem.
Both Matching networks and SNAIL are trained using the multi-task training loss and applied to test tasks without any finetuning.
\cutspacepara

\paragraph{MAML~\citep{finn2017model}} 
All model parameters are trained using MAML, with the same model architecture as TAM.
The entire model is fine-tuned on test tasks.
\cutspacepara

\paragraph{CAVIA~\citep{zintgraf2018caml}} 
Similar to TAM, CAVIA has a set of task-specific parameters and shared parameters.
The training algorithm is similar to MAML, but inner loop updates are performed on the task-specific parameters as opposed to the entire model.

\subsection{Sequence Classification and Transduction} \label{sec:exp_noncomp}
\cutspacebelow

Most prior work on few-shot learning have focused on computer vision benchmarks such as Omniglot \citep{lake2019compositional} and Mini-ImageNet \citep{vinyals2016matching}.
In the sequential data setting, \citet{bao2019few} constructed synthetic benchmarks from existing text datasets but the number of tasks is rather limited. 
In this work we construct a new set of benchmarks involving synthetic sequential data, allowing us to evaluate models in a more controlled setting after training on a larger number of tasks.

\subsubsection{Synthetic Benchmarks}
\label{sec:benchmarks}
\cutspacebelow

We construct a synthetic few-shot {\em classification} benchmark as follows. 
The benchmark consists of tasks that involve a non-negative integer sequence as input and a discrete label as output. 
A task is constructed by applying a sequence of mathematical transformations to input sequences as follows:
Element-wise transform ($T_1$) {$\rightarrow$} Subsequence extraction ($T_2$) {$\rightarrow$} Labeling function $(T_3)$. 
The arrows indicate function composition and the sequence of transformations maps an input sequence to a single integer. 
The transformations are defined as 
$T_1 \in S_1, T_2 \in S_2, T_3 \in S_3$ where, 
\textit{
$S_1 = $ \{mul $v$, add $v$, div $v$, mod $v$\}; 
$S_2 = $ \{(not) multiple of $v$, (not) greater than $v$, (do not) have exactly $v$ divisors\};
$S_3 = $ \{count, min, max, mean, median, mode, first, last, max-min, middle\}}, where $v \in \{1\cdots n\}$ for some integer $n$.
We randomly generate a large number of sequences $X$ of integers from $\{0 \cdots N\}$.
We apply the transformation sequence $T_1, T_2, T_3$ to these sequences $x \in X$ and get the corresponding outputs $T_3(T_2(T_1(x)))$.
The $C$ most frequent outputs are then defined to be the $C$ classes of interest. 
We obtain a uniform amount of data from each class and discard input sequences for which the output does not belong to one of the chosen $C$ classes.  
Cases where at least $C$ distinct outputs cannot be obtained are discarded.
These $C$ classes then constitute a $C$-way classification task.
An example task is {\tt mul 2 $\rightarrow$ less than 5 $\rightarrow$ count}, where the goal is to count the number of input elements which, when multiplied by 2, are less than 5 (i.e., count number of input integers less than 3). 
The semantics of each of the transforms are defined in the appendix.
We set $C = 4$ in our experiments. 
Vocabulary size and input sequence length are set to 12 and 5, respectively. 
Combinations of transforms that have identical input-output relationship are identified and removed during task construction.
All tasks are thus unique in terms of input-output mapping.

We also construct two sequence transduction benchmarks.
The first benchmark is constructed in a way similar to the classification tasks where we consider a sequence of transformations mapping an input sequence to an output sequence $T_1 \rightarrow T_2 \rightarrow T_3$, where 
$T_1 \in S_1, T_2 \in S_2, T_3 \in S_3$;
\textit{
$S_1 = $ \{mul $v$, add $v$, div $v$, mod $v$\}; 
$S_2 = $ \{replace $v$ with $v'$, replace $x_i$ with $f(x_i, x_j)$ \}; 
$S_3 = $ \{sort ascending, sort descending, reverse, swap$(x_i, x_j)$, shift right $v$\}}, 
and $v, v', i, j$ are integers chosen at random, $x_p$ represents the element at position $p$ in the input sequence, $f$ is a mathematical function (Eg: $f(a, b) \in \{a + b, \text{abs}(a-b), b, \cdots\}$).
An example task is {\tt add 2 $\rightarrow$ replace 2 with 1 $\rightarrow$ reverse}, and an (input, output) sample drawn from this task is: $([0,5,0,3,6], [8,5,1,7,1])$.

Our second transduction benchmark is a path finding task in a grid world (see appendix \ref{sec:pathfinding} for an illustration).
A task is defined by start and end positions in a square grid of size $N \times N$.
Given the locations of obstacles in this grid, the objective of the task is to find the shortest path connecting start and end positions that avoids the obstacles. 
The source and target sequences correspond to the locations of obstacles and optimal path from start to end position avoiding the obstacles, respectively. 

We use 500, 16, 64 tasks respectively for training, validation and testing for all three setups.
Tasks are unique and randomly assigned to these sets, in other words we test generalization under the condition of distributional match between the training and the test set. 
Each training task has 500 examples.

\subsubsection{Results}
\cutspacebelow
\input{results/main_results.tex}

Table~\ref{tab:classification} reports the results on this benchmark.
In the extreme few-shot setting ($k=1$), all methods perform poorly, although memory based methods such as matching networks and SNAIL fare the best. However, they start performing relatively worse when more labelled data is available, where fine-tuning part or all of the model parameters could be beneficial.   Both SNAIL and matching networks sometimes perform {\it absolutely} worse when more labeled examples are present, suggesting they are failing to effectively use their memory when confronted with longer sequences. %
Fine-tuning the whole model, particularly in the multitask setting, works remarkably well for larger values of $k$, although the best performance is achieved by TAM, suggesting the need for sample efficient task adaptation methods.
For $k>1$, TAM performs comparably or better than all baselines, including MAML and CAVIA.  
Furthermore, TAM is more efficient to train than CAVIA (see section \ref{sec:discussion}).

\subsection{Compositional Task Representations}
\cutspacebelow
\label{sec:comp_exp}

Compositional reasoning is arguably an important skill for few-shot learning~\citep{lake2019compositional,purushwalkam2019task}.
The underlying assumption is that there exist primitive skills which can be learned and combined together to solve entirely new tasks. 
If a learner can leverage the compositional structure of the learning task, it may learn with even less labeled data.

In this section we assess how much better TAM works when we expose the compositional structure of the tasks described in 
\textsection\ref{sec:benchmarks}.
Specifically, we assess the ability to learn new tasks which are composed of primitives, some of which were unseen during training.
To present an example from the classification setting, assume the models know that tasks are composed of three transforms $T_1\in S_1, T_2\in S_2, T_3\in S_3$.
We henceforth refer to the elements of $S_1 \cup S_2 \cup S_3$ as \textit{primitives}. 
Further assume the model never saw the \textit{add $k$} primitive during training.
Given a new test task for which $T_1 = $\textit{ add }3 (and $T_2, T_3$ are known primitives seen during training), we expect the model to infer the concept of \textit{add} from the few training examples of the test task.

\subsubsection{Task Construction}
In the compositional setting, we provide models with information about the primitives used to construct the task.
For the classification and transduction tasks, the training and test tasks are constructed as follows. 
Assume the set of primitives available for the three transforms to be $S_1, S_2, S_3$.
We hold out a subset of primitives $S_1', S_2', S_3'$ respectively from each of these three sets, which shall constitute the \textit{unseen} primitives.
The training tasks are made up of primitives from $S_1 - S_1', S_2 - S_2', S_3 - S_3'$, which we will refer to as \textit{seen} primitives.
The test tasks are made up of seen and unseen primitives where exactly one primitive is unseen (For instance, $T_1 \in S_1 - S_1', T_2 \in S_2', T_3 \in S_3 - S_3'$). %
Model performance is averaged over multiple (8) different choices of $S_1', S_2', S_3'$.

\input{results/compositional.tex}

We also define a compositional path-finding task as follows.
In addition to finding the optimal path from start, end positions while avoiding obstacles, we now require the path to lie on a specified way-point. %
The locations of the start, end and way points thus define the primitives that make up a task.
Similar to the previous settings, we hold out sets of values for each of these points and construct the train/test tasks in an analogous manner.

\subsubsection{Training}

For all the models, a sequence of primitive ids representing the primitives that make up the task is appended to the input sequence.
These primitive embeddings $\theta_e$ are learned along with the other model parameters.
To simulate the testing conditions, at training time we pretend some primitives are unknown.
For the multitask, matching network and SNAIL baselines we learn an \textit{unknown primitive embedding}, which is used to initialize embeddings of unknown primitives encountered at test time.
 Although the tasks themselves are harder (because entire primitives are unseen), modeling them is easier because the primitive information is given to the model. 
For CAVIA and TAM, we infer embeddings for unknown primitives on the fly using gradient descent during train and test.
See Algorithm \ref{alg:comptraining} in the appendix for the complete algorithm. 
We use 5000 training tasks, and 100 validation and test tasks each. 
Each training task has 500 examples.

\subsubsection{Results}
Table~\ref{tab:comp} summarizes the results in the compositional setting. We observe similar trends as before for the non-compositional case. Multitask learning becomes competitive only for larger values of $k$. Vice versa, matching networks and SNAIL suffer with long sequences (larger values of $k$). TAM performs at least comparably if not better than methods relying on second order derivatives like CAVIA. In fact, CAVIA sometimes fail to converge as shown by the rather large error bars. Finally, the compositional version of TAM often yields higher accuracy than the corresponding non-compositional version, showing that the model is able to cleverly leverage the additional knowledge about a subset of primitives (two out of three) that compose the new task.
Compositionality is particularly helpful with fewer shots (e.g., 1-shot) -- with sufficient training examples (e.g., 20-shot) models benefit less from compositionality.

\subsection{Ablation experiments}
\cutspacebelow
\input{results/ablation.tex}

\paragraph{Where to plug task embeddings}
The experiments in the paper so far consider a simple conditioning scheme where the task embedding appears as an additional embedding in the input sequence of token embeddings. 
We compare this against other ways of incorporating task-specific parameters into the model.
\citet{houlsby2019parameter} introduce adapter layers, parameter modules that are inserted at every layer of a pre-trained transformer. 
An adapter layer down-projects its input, applies a non-linearity, and up-projects the representation back to the original size.
Our first baseline considers parameters in the adapter layers as the task embedding.
Another popular method for adapting pre-trained networks to new tasks is adapting parameters in normalization layers \citep{perez2018film,ghiasi2017exploring}. 
In our second baseline, we consider the scale and bias parameters in the Layer Normalization layers of the transformer as the task embedding.

The results in Table~\ref{tab:ablation} on non-compositional classification tasks show that using adapter layer parameters as task embedding yields similar results to the simplest conditioning scheme where the task embedding is fed as an additional input. Using normalization parameters as the task embedding instead performs slightly worse. 
This shows that our input conditioning scheme is simple yet effective.

\paragraph{Importance of transformer architecture}

The experiments presented in this paper so far have used a transformer architecture.
Although transformers are a natural choice for problems involving sequences owing to their recent success, the proposed training algorithm applies equally well to other architectures.
We study the impact of swapping out the transformer with a recurrent model in Table \ref{tab:achitecture} on non-compositional tasks.
We use a bidirectional LSTM with a comparable number of parameters to our transformer model. 
The classifier head acts on the final representation of the final layer of the LSTM.
We examine the performance of the two architectures when trained using both multitasking and the proposed alternating minimization training algorithm.
First, we observe that the transformer generally performs better than the recurrent model.
Second, the proposed training algorithm yields consistent improvements over the multitask baseline for the transformer.
This shows that the proposed algorithm is general, but particularly effective when used in conjunction with the transformer architecture.

\paragraph{Visualizing learned task embeddings}

In Figure \ref{fig:embedding_map} we visualize task embeddings learned by the non-compositional TAM model in our gridworld task.
We visualize the first two principal components of task embeddings corresponding to tasks which have the same start position. %
The projections are color coded by the horizontal and vertical coordinates of the end position for each task.
This shows that the task embeddings have learned the structure of the tasks.

\input{results/timing_tsne.tex}

\subsection{Discussion}
\label{sec:discussion}

\paragraph{Optimizing Task Embeddings} 
We observed that both CAVIA and TAM generally attain better performance when trained with a  larger number of inner loop updates. In this work, we use a maximum of 25 inner loop updates for TAM because it strikes a good balance between finding an optimal task embedding and containing training time.
CAVIA performed best with 10 inner loop updates, beyond which we hit the computational limitations  of our hardware. 
We also found that TAM works better when trained with a number of examples per task much greater than $k$, in our case 300. All these empirical findings suggest that optimizing for the task embedding and replacing the second order optimization with TAM's first order is an intrinsically  difficult problem that requires more iterations and a larger number of examples. 

\paragraph{Training Efficiency}
We discuss the training efficiency of different models in Table \ref{tab:timing}.
The multitask baseline is not expensive to train, but it doesn't perform well on few-shot scenarios.
CAVIA does well especially in the extreme few-shot scenarios, but has stability issues.
TAM is simple, easy to implement, performs comparably or better than the baselines and trains more efficiently than CAVIA.

\paragraph{First vs. Second Order Gradients}
Double backprop to optimize the test optimization has become a standard method of meta-learning.  
In the appendix of \citet{finn2017model}, and in \citet{zintgraf2018caml} (the ``first order'' variant), similar approaches to TAM were shown to perform relatively worse than the methods with second order gradients.
In contrast, in our settings, we have found that first order gradients (via alternating minimization) are sufficient if done correctly, despite being simpler and more efficient.  
Although CAVIA~\citep{zintgraf2018caml} sometimes outperforms TAM, especially for very small numbers of test examples,  TAM is always competitive; with more test examples, TAM is usually superior.  
TAM always outperforms MAML~\citep{finn2017model}.

%% file: results/main_results.tex
\setlength{\tabcolsep}{5pt}
\begin{table*}[!t]
\small
\centering
\begin{tabular}{l | c c c c | c c c c | c c c c}
\multirow{2}{*}{Model}& \multicolumn{4}{c}{Sequence Classification} & \multicolumn{4}{c}{Sequence Transduction} & \multicolumn{4}{c}{Path Finding} \\
& 1 & 5 & 10 & 20 & 1 & 5 & 10 & 20 & 1 & 5 & 10 & 20 \\
\midrule
Task & 40.50 & 64.75 & 74.25	& 82.50 & 6.27 & 5.26 & 4.71 & 4.01 & 3.17 & 1.75 & 1.55 & 1.39  \\[-0.2em]
Agnostic & \std 1.73 & \std 1.29 & \std 1.29 & \std 1.58 & \std 0.17 & \std 0.04 & \std 0.05 & \std 0.07 & \std 0.27 & \std 0.03 & \std 0.02 & \std 0.01 \\[0.3em]
Multitask & 38.75 & 66.00	& 77.50	& 87.50 & 13.80 & 6.80 & 5.18 & 2.91 & 6.39 & 1.98 & 1.64 & 1.44 \\[-0.2em]
& \std 0.96 & \std 0.82 & \std 1.29 & \std 0.58 & \std 3.36 & \std 0.26 & \std 1.01 & \std 0.49 & \std 1.96 & \std 0.12 & \std 0.05 & \std 0.02 \\
\midrule 
Matching & \textbf{43.00} & 58.75 & 64.50 & 67.00 & \multicolumn{4}{c|}{\multirow{2}{*}{--}} & \multicolumn{4}{c}{\multirow{2}{*}{--}} \\[-0.2em]
Network & \std 1.15 & \std 2.45 & \std 1.83 & \std 1.41 & \multicolumn{4}{c|}{} & \multicolumn{4}{c}{} \\[0.3em]
SNAIL & \textbf{43.00} & 44.00	& 68.25	& 67.50 & \textbf{2.48} & 3.80 & 4.98 & 4.11 & 2.63 & 1.95 & 3.47 & 3.04 \\[-0.2em]
& \std 1.41 & \std 2.00 & \std 1.26 & \std 4.43 & \std 0.38 & \std 0.38 & \std 0.03 & \std 2.94 & \std 0.27 & \std 0.25 & \std 1.56 & \std 1.01 \\
\midrule
MAML & 39.60 & 63.40 & 71.80 & 78.80 & 6.73 & 5.84 & 5.19 & 4.20 & 5.49 & 2.05 & 1.65 & 1.44 \\[-0.2em]
& \std 0.55 & \std 0.89 & \std 0.84 & \std 0.84 & \std 0.16 & \std 0.2 & \std 0.08 & \std 0.10 & \std 0.85 & \std 0.03 & \std 0.01 & \std 0.01 \\[0.3em]
CAVIA & \textbf{43.00} & \textbf{78.00}	& 87.00	& 91.00 & 11.05 & \textbf{2.75}  & 1.78 & 1.53 & 2.21 & 1.31 & 1.25 & 1.21 \\[-0.2em]
& \std 0.58 & \std 1.26 & \std 0.50 & \std 0.58 & \std 2.94 & \std 0.51 & \std 0.14 & \std 0.06 & \std 0.21 & \std 0.03 & \std 0.03 & \std 0.02 \\[0.3em]
TAM & 40.50 & 75.50	& \textbf{89.50} & \textbf{94.50} & 8.47 & 2.92 & \textbf{1.47} & \textbf{1.15} & \textbf{1.82} & \textbf{1.27} & \textbf{1.22} & \textbf{1.17} \\[-0.2em]
& \std 0.82 & \std 0.50 & \std 0.58 & \std 0.82 & \std 1.42 & \std 0.67 & \std 0.18 & \std 0.03 & \std 0.08 & \std 0.01 & \std 0.01 & \std 0.01 \\
\end{tabular}
\caption{$k$-shot sequence classification and sequence transduction experiments on our three benchmarks for $k \in \{1, 5, 10, 20\}$.
The metric for sequence classification is average accuracy on test tasks (higher is better).
On the transduction tasks, the performance metric is average perplexity on test tasks (lower is better). Random performance is at 25\% accuracy (classification) and 12 perplexity points (other two tasks). Entries in smaller font are error bars, and they are estimated on 4 trials varying the model initialization.
}
\label{tab:classification}
\cuttablebelow
\end{table*}

%% file: results/compositional.tex
\setlength{\tabcolsep}{4pt}
\begin{table*}[!t]
\small
\centering
\begin{tabular}{l | c c c c | c c c c | c c c c}
\multirow{2}{*}{Model} & \multicolumn{4}{c}{Sequence Classification} & \multicolumn{4}{c}{Sequence Transduction} & \multicolumn{4}{c}{Path Finding} \\
& 1 & 5 & 10 & 20 & 1 & 5 & 10 & 20 & 1 & 5 & 10 & 20 \\
\midrule
Multitask & 57.50 & 74.00 & 81.00 & 88.5 & 43.32 & 7.50 & 3.48 & 2.16 & 3.08 & 1.62 & 1.32 & 1.23 \\[-0.2em]
& \std 3.51 & \std 5.48 & \std 4.24 & \std 2.08 & \std 10.87 & \std 0.43 & \std 0.12 & \std 0.05 & \std 0.61 & \std 0.33 & \std 0.05 & \std 0.02 \\
\midrule
Matching & 61.25 & 69.50 & 72.25 & 67.5 & \multicolumn{4}{c|}{\multirow{2}{*}{--}} & \multicolumn{4}{c}{\multirow{2}{*}{--}} \\[-0.2em]
Network & \std 4.35 & \std 5.45 & \std 6.08 & \std 5.92 & \multicolumn{4}{c|}{} & \multicolumn{4}{c}{} \\[0.3em]
SNAIL & \textbf{63.5} & 71.75 & 76.25 & 80.25 & 7.00 & 6.24 & 6.93 & 17.10 & \textbf{1.53} & 1.71 & 3.36 & 4.22 \\[-0.2em]
& \std 4.80 & \std 4.86 & \std 2.75 & \std 2.87 & \std 2.03 & \std 0.27 & \std 3.25 & \std 9.66 & \std 0.29 & \std 0.09 & \std 0.57 & \std 0.79 \\
\midrule
CAVIA & 57.25 & 66.25 & 67.50 & 68.50 & 36.72 & 6.01 & 3.99 & 3.27 & 1.99 & 1.27 & 1.21 & 1.17 \\[-0.2em]
 & \std 12.09 & \std 14.73 & \std 15.67 & \std 16.42 & \std 8.83 & \std 0.88 & \std 0.50 & \std 0.24 & \std 0.11 & \std 0.01 & \std 0.00 & \std 0.00 \\[0.3em]
TAM & 63.00 & \textbf{76.50} & \textbf{82.75} & 88.5 & \textbf{6.15} & \textbf{3.43}  & \textbf{2.69}  & \textbf{2.13}  & 2.33  & 1.30  & 1.23  & 1.19  \\[-0.2em]
(Comp) & \std 5.35 & \std 4.65 & \std 3.86 & \std 2.65 & \std 0.91 & \std 0.05 & \std 0.04 & \std 0.02 & \std 0.18 & \std 0.02 & \std 0.01 & \std 0.01 \\
\midrule 
TAM & 45.25 & 72.5 & 81.5 & \textbf{89.75} & 7.80 & 5.08 & 3.60 & 2.42 & 4.37 & \textbf{1.27} & \textbf{1.17} & \textbf{1.11} \\[-0.2em]
(Non-comp) & \std 3.59 & \std 3.70 & \std 2.89 & \std 0.96 & \std 0.09 & \std 0.24 & \std 0.15 & \std 0.08 & \std 3.59 & \std 0.01 & \std 0.00 & \std 0.00 \\

\end{tabular}
\caption{Compositional models for few-shot sequence classification and sequence transduction.  All models (except non-compositional TAM) get information on the primitives present in the tasks via extra tokens appended to the input sequence, except that one such primitive is unseen at test time. Non-compositional TAM is not given information about primitives, and estimates a single task embedding instead.}
\label{tab:comp}
\cuttablebelow
\end{table*}

%% file: results/ablation.tex
\begin{figure}[!t]
{
\small
\begin{minipage}[t]{.45\textwidth}
\centering
\begin{tabular}{l | c c c c}
\multirow{2}{*}{Model}& \multicolumn{4}{c}{Sequence Classification (Accuracy)} \\
 & 1-shot & 5-shot & 10-shot & 20-shot \\
\midrule
Input token             & 0.41 & \textbf{0.76} & \textbf{0.89} & \textbf{0.94} \\
Adapters                & \textbf{0.43} & 0.75 & 0.88 & \textbf{0.94} \\
Layer Norm              & 0.42 & 0.64 & 0.78 & 0.88 \\
\end{tabular}
\captionof{table}{$k$-shot classification accuracy when plugging the task embedding in various ways for different values of $k$.}
\label{tab:ablation}
\end{minipage}\qquad
\begin{minipage}[t]{.50\textwidth}
\centering
\begin{tabular}{l | c | c c c c }
\multirow{2}{*}{Arch} & \multirow{2}{*}{Training} & \multicolumn{4}{c}{Sequence Classification (Acc.) } \\
& & 1-shot & 5-shot & 10-shot & 20-shot \\
\midrule 
\multirow{2}{*}{LSTM} & Multi & 0.38 & 0.57 & 0.72 & 0.87 \\
& Alt & 0.35 & \textbf{0.78} & 0.83 & 0.85 \\
\midrule
\multirow{2}{*}{Transf.} & Multi & 0.39 & 0.68 & 0.80 & 0.88 \\
& Alt & \textbf{0.41} & 0.76 & \textbf{0.89} & \textbf{0.94} \\
\end{tabular}
\captionof{table}{k-shot accuracy for different architectures with multitask and the proposed training algorithms.}
\label{tab:achitecture}

\end{minipage}
}
\cuttablebelow
\end{figure}

%% file: results/timing_tsne.tex
\begin{figure}[!t]
{
\small
\begin{minipage}[b]{.45\textwidth}
\centering
\includegraphics[width=0.45\linewidth, trim={2cm 2cm 2cm 2cm},clip]{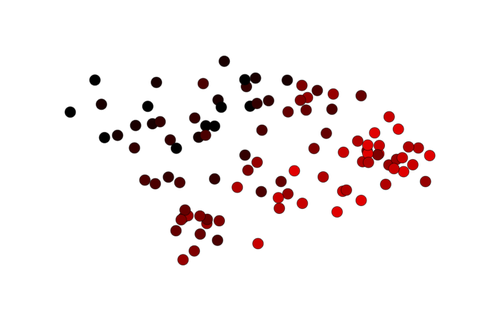}
\includegraphics[width=0.45\linewidth, trim={2cm 2cm 2cm 2cm},clip]{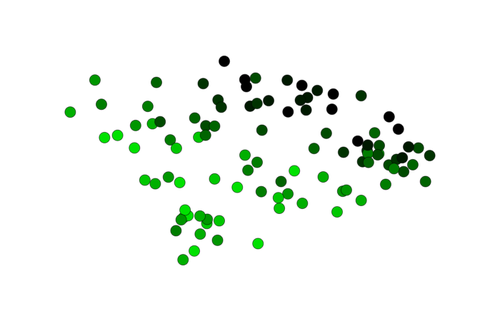}
\captionof{figure}{
2D PCA projections of task embeddings learned by our algorithm for the gridworld domain. 
Tasks visualized here have the same start position (4,4).
Points are color coded based on horizontal (left plot) and vertical (right plot) coordinates of the end position corresponding to each task.
}
\label{fig:embedding_map}
\end{minipage}\quad
\begin{minipage}[b]{.50\textwidth}
\centering
\begin{tabular}{l | l l l}
\multirow{2}{*}{Model} & Classification & Transduction & Path-finding \\
& Acc., Time & Ppl., Time & Ppl.,  Time \\
\midrule
Multitask   & 67.4, \textbf{30min} & 7.2, \textbf{23min}  & 2.9, \textbf{20min} \\
CAVIA       & 74.8, 3h   &   4.5, 5.3h   & 1.5, 3.7h \\
TAM         & \textbf{75.0}, 2h   & \textbf{1.5}, 2.3h   & \textbf{1.3}, 2.7h
\end{tabular}
\captionof{table}{Training efficiency: Time taken by each training algorithm to reach the best model (identified using validation tasks) and corresponding model performance (non-compositional setting). Performance and time are averaged across {\scriptsize $k\in\{1,5,10,20\}$} shots.}
\label{tab:timing}
\end{minipage}
}
\cuttablebelow
\end{figure}

%% file: sections/conclusion.tex
\cutspaceabove
\section{Conclusion}
\cutspacebelow
In this work we demonstrate a simple and effective approach to adapt transformer models to new tasks with limited data.
TAM is trained to adapt to new tasks by inferring a small set of parameters called the \textit{task embedding} using gradient descent.
On synthetic sequence classification and transduction benchmarks we constructed TAM yields comparable or superior performance to approaches relying on second order derivatives, while being computationally more efficient.

%% file: sections/supplement.tex
\section{Sequence transformations used to construct classification and transduction tasks}

In tables \ref{tab:clstf}, \ref{tab:transtf} we describe the transformations used to construct classification and transduction tasks, respectively.

\begin{table*}[!ht]
\centering 
\begin{tabular}{c p{3.5cm} p{9cm}}
\toprule
& \textbf{Transformation} & \textbf{Description} \\
\midrule 
\multirow{4}{*}{$S_1$} & mul $v$ & Elementwise multiply by $v$ \\
& add $v$ & Elementwise add $v$ \\
& div $v$ & Elementwise integer division by $v$ \\
& mod $v$ & Elementwise modulo $v$ operation \\
\midrule 
\multirow{3}{*}{$S_2$} & (not) multiple of $v$ & Extract subset of integers that are (not) multiples of $v$ \\
& (not) greater of $v$ & Extract subset of integers that are (not) greater than $v$ \\
& (do not) have exactly $v$ divisors & Extract subset of integers that (do not) have exactly $v$ divisors \\
\midrule 
\multirow{10}{*}{$S_3$} & 
count & Sequence length \\
& min & Smallest integer in sequence \\
& max & Largest integer in sequence \\
& mean & Mean of sequence elements \\
& median & Median of sequence elements \\
& mode & Mode of sequence elements \\
& first & First element in sequence \\
& last & Last element in sequence \\
& max-min & Difference between largest and smallest elements in sequence \\
& middle & Element in the middle position of sequence \\
\bottomrule
\end{tabular}
\caption{Sequence transformations used to construct classification tasks and their descriptions. Each transformation takes a sequence as input and outputs a sequence (transformations in $S_1$ and $S_2$), or a single integer (transformations in $S_1$).}
\label{tab:clstf}
\end{table*}

\begin{table*}[!ht]
\centering 
\begin{tabular}{c p{3.5cm} p{9cm}}
\toprule
& \textbf{Transformation} & \textbf{Description} \\
\midrule 
\multirow{3}{*}{$S_1$} & mul $v$ & Elementwise multiply by $v$ \\
& add $v$ & Elementwise add $v$ \\
& div $v$ & Elementwise integer division by $v$ \\
& mod $v$ & Elementwise modulo $v$ operation \\
\midrule 
\multirow{3}{*}{$S_2$} & reverse $v$ with $v'$ & Replace all occurrences of $v$ in the sequence with $v'$ \\
& replace $x_i$ with $f(x_i, x_j)$ & Replace element $x_i$ with one of the following: \\
& & \{$ax_i+b$, $x_j$, abs$(x_i-x_j)$, $x_i + x_j$\} where $a, b$ are integer constants and $x_i, x_j$ are elements of the sequence at position $i, j$ respectively \\
\midrule 
\multirow{3}{*}{$S_3$} & sort ascending & Sort the sequence in ascending order \\
& sort descending & Sort the sequence in descending order \\
& reverse & Reverse the sequence \\
& swap$(x_i, x_j)$ & Swap elements at positions $i, j$ of the sequence\\
& shift right $v$ & Cyclic shift the sequence right by $v$ positions \\
\bottomrule
\end{tabular}
\caption{Sequence transformations used to construct transduction tasks and their descriptions. Each transformation takes a sequence as input and outputs a sequence. }
\label{tab:transtf}
\end{table*}

\newpage 

\section{Path-finding task}
\label{sec:pathfinding}

\subsection{Non-compositional path-finding task}
We present an example task from the path-finding task below.
The grids are $10\times 10$.
The following task is defined by the start position (7, 0) and end position (1, 4), indicated by the green and red squares, respectively.
Each example in the task corresponds to a particular configuration of obstacles in the grid. 
The source sequence represents the locations of obstacles.
The obstacles are represented by the top left position of a $2\times 2$ blob.
The target sequence represents the optimal path from source to target.
Source and target sequences consist of rasterized grid coordinates (Eg. rasterized start and end positions are 70 and 14, respectively).
In addition, elements of the target sequence have an offset of 100 (Eg. rasterized position 14 is represented as 114).

\def\Side{\ChessSide}
\newcommand\ChessBoxA{%
  {\fboxsep=0pt\fbox{\color{\ChessColori}\rule{\Side}{\Side}}}}
\newcommand\ChessBoxB{%
  {\fboxsep=0pt\fbox{\color{\ChessColorii}\rule{\Side}{\Side}}}}
\newcommand\ChessBoxS{%
  {\fboxsep=0pt\fbox{\color{\ChessColoriii}\rule{\Side}{\Side}}}}
\newcommand\ChessBoxT{%
  {\fboxsep=0pt\fbox{\color{\ChessColoriv}\rule{\Side}{\Side}}}}
\newcommand\ChessBoxW{%
  {\fboxsep=0pt\fbox{\color{\ChessColorv}\rule{\Side}{\Side}}}}

\makeatletter
\newcommand\Row[1]{%
  \par\nobreak\nointerlineskip\vskip-\fboxrule%
  \@tfor\@tempa:=#1 \do {\csname ChessBox\@tempa\endcsname\kern-\fboxrule}}
\define@key{chessB}{side}{\def\ChessSide{#1}}
\define@key{chessB}{colori}{\def\ChessColori{#1}}
\define@key{chessB}{colorii}{\def\ChessColorii{#1}}
\define@key{chessB}{coloriii}{\def\ChessColoriii{#1}}
\define@key{chessB}{coloriv}{\def\ChessColoriv{#1}}
\define@key{chessB}{colorv}{\def\ChessColorv{#1}}
\setkeys{chessB}{
  side=0.5em,
  colori=white,
  colorii=black,
  coloriii=green,
  coloriv=red,
  colorv=yellow}
\makeatother

\newenvironment{Chessboard}[1][]
  {\setkeys{chessB}{#1}%
  \par\medskip\setlength\parindent{0pt}}
  {\par\medskip}

\begin{center}
\begin{minipage}{6em}
\begin{Chessboard}
\Row{A,A,A,A,A,A,A,A,A,B}
\Row{A,A,A,A,T,A,A,A,A,B}
\Row{A,A,A,A,A,A,A,A,A,A}
\Row{A,B,B,A,A,A,A,A,A,B}
\Row{A,B,B,A,B,B,A,A,A,B}
\Row{A,B,B,A,B,B,A,A,A,A}
\Row{A,B,B,B,B,A,A,A,A,B}
\Row{S,A,A,B,B,A,A,A,B,B}
\Row{A,A,A,A,A,A,A,A,B,B}
\Row{A,A,A,A,A,A,A,A,A,A}
\end{Chessboard}
\end{minipage}
\begin{minipage}{6em}
\begin{Chessboard}
\Row{A,A,A,A,A,A,A,A,A,A}
\Row{A,A,B,B,T,A,A,A,A,A}
\Row{A,B,B,B,A,B,B,A,A,A}
\Row{A,B,B,B,A,B,B,A,A,A}
\Row{A,A,A,A,B,B,B,A,A,A}
\Row{A,A,A,A,B,B,A,A,A,A}
\Row{A,A,B,B,A,A,A,A,A,A}
\Row{S,A,B,B,A,A,A,A,A,A}
\Row{A,A,A,A,A,A,A,A,A,A}
\Row{A,A,A,A,A,A,A,A,A,B}
\end{Chessboard}
\end{minipage}
\begin{minipage}{6em}
\begin{Chessboard}
\Row{A,B,B,A,A,A,A,A,A,A}
\Row{A,B,B,A,T,A,A,A,A,A}
\Row{A,A,A,A,A,A,A,A,A,A}
\Row{A,B,B,A,B,B,A,A,A,A}
\Row{A,B,B,A,B,B,A,A,A,A}
\Row{B,B,A,A,A,A,A,A,A,A}
\Row{B,B,A,A,A,A,A,A,A,A}
\Row{S,A,A,A,A,A,A,A,A,A}
\Row{A,A,A,A,A,A,A,A,A,A}
\Row{B,B,A,A,B,B,B,B,A,B}
\end{Chessboard}
\end{minipage}
\end{center}

\begin{itemize}[leftmargin=*]
\item Source: [39, 78, 51, 9, 31, 63, 44, 69], Target: [170, 160, 150, 140, 130, 121, 112, 103, 114]
\item Source: [12, 35, 99, 22, 62, 44, 25, 21],  Target: [170, 161, 152, 143, 134, 124, 114]
\item Source: [90, 99, 1, 96, 34, 50, 94, 31],  Target: [170, 171, 162, 152, 143, 133, 123, 114]
\end{itemize}

\subsection{Compositional path-finding task}
\label{sec:comppathfinding}

In the compositional setting, we require the optimal path to pass through a way-point, indicated in yellow in the following grids.
A task is thus defined by a start position, end position and way-point position.
The possible values for each of these three parameters represent the primitives in this compositional setting.

\begin{center}
\begin{minipage}{6em}
\begin{Chessboard}
\Row{A,A,A,A,A,A,A,A,A,A}
\Row{A,A,A,A,A,A,A,A,B,B}
\Row{A,A,T,A,A,A,S,A,B,B}
\Row{A,A,A,A,A,A,A,A,B,B}
\Row{A,A,W,A,A,A,A,A,B,B}
\Row{A,A,A,A,A,A,A,A,A,A}
\Row{A,A,A,B,B,A,A,B,B,B}
\Row{A,A,A,B,B,A,A,B,B,B}
\Row{A,A,A,B,B,A,A,A,A,A}
\Row{B,B,A,B,B,A,A,A,A,A}
\end{Chessboard}
\end{minipage}
\begin{minipage}{6em}
\begin{Chessboard}
\Row{A,A,A,A,A,A,A,A,A,A}
\Row{A,A,A,A,A,A,A,A,A,A}
\Row{A,A,T,B,B,A,S,A,A,A}
\Row{A,A,A,B,B,B,A,A,A,A}
\Row{A,A,W,A,B,B,A,A,A,A}
\Row{A,A,A,A,A,A,A,A,A,A}
\Row{B,B,A,A,A,A,A,A,A,A}
\Row{B,B,A,A,A,A,A,A,A,A}
\Row{B,B,A,B,B,B,B,A,A,A}
\Row{A,A,A,B,B,B,B,A,A,A}
\end{Chessboard}
\end{minipage}
\begin{minipage}{6em}
\begin{Chessboard}
\Row{A,A,A,A,A,A,A,A,B,B}
\Row{A,A,A,B,B,A,A,A,B,B}
\Row{A,A,T,B,B,A,S,A,A,B}
\Row{A,A,A,A,A,A,A,A,A,B}
\Row{A,A,W,A,A,A,A,A,A,A}
\Row{A,A,A,B,B,A,A,B,B,A}
\Row{A,A,A,B,B,A,A,B,B,A}
\Row{A,A,A,A,A,A,A,B,B,A}
\Row{A,A,A,A,A,A,A,B,B,A}
\Row{A,B,B,A,A,A,B,B,A,A}
\end{Chessboard}
\end{minipage}
\end{center}

\begin{itemize}[leftmargin=*]
\item Source: [63, 38, 90, 93, 73, 68, 18, 67], Target: [126, 115, 124, 133, 142, 131, 122]
\item Source: [95, 60, 95, 70, 23, 34, 83, 85], Target:  [126, 115, 104, 113, 122, 131, 142, 131, 122]
\item Source: [91, 29, 57, 96, 8, 53, 77, 13], Target: [126, 125, 134, 133, 142, 131, 122]
\end{itemize}

\newpage
\section{Compositional TAM}

Algorithm \ref{alg:comptraining} presents the training algorithm for compositional TAM.
We draw a training task $\mathcal{T}^\text{train}$ with primitive ids $T_1 = i_1, T_2 = i_2, T_3 = i_3$ respectively in line 3.
These primitive ids index into the primitive embedding table $\theta_e$. 
We pretend that one of the primitives is unknown, and to illustrate the algorithm, we assume without loss of generality that $T_2=i_2$ is unknown (line 5).
In the inner loop optimization, we infer an embedding $z$ for this unknown primitive using gradient descent, while using the primitive embedding table to load the known primitive embeddings ($\theta_e[i_1], \theta_e[i_3]$ in this case (lines 8, 9)). 

\begin{algorithm*}[!hb]
    \setcounter{AlgoLine}{0}
    \SetKwInOut{Input}{Input}
    \SetKwInOut{Output}{Output}

    \Input{Training tasks $\mathcal{T}_1^\text{train}, ..., \mathcal{T}_N^\text{train}$}
    \Output{Model parameters $\theta$, primitive embeddings $\theta_e$}
    
    $\theta' = \theta \cup \theta_e$

    \Repeat{max training iterations}{
      Sample training task $\mathcal{T}^\text{train}$\, with primitive ids $T_1 = i_1, T_2 = i_2, T_3 = i_3$
      
      Sample $k$ training examples from the task 
      $\{(x^j,y^j)_{j=1,\cdots,k}\} \sim \mathcal{T}^\text{train}$

      Pretend one of the primitives (chosen at random) is unknown, say $T_2$

      Initialize $z = 0, \Delta \theta' = 0$

      \While{loss improves and max iterations not reached }
      {
        $z \leftarrow z - \nabla_{z} \sum_{j=1}^k  \, - \log p(y^j | x^j, z_1 = \theta_e[i_1], z_2 = z, z_3 = \theta_e[i_3]; \theta')$
        
        $\Delta \theta' \leftarrow \Delta \theta' - \nabla_{\theta'} \sum_{j=1}^k - \log p(y^j | x^j, z_1 = \theta_e[i_1], z_2 = z, z_3 = \theta_e[i_3]; \theta') $
      }
      $\theta' \leftarrow \theta' + \Delta \theta'$
      
    }
    \caption{\textbf{Compositional TAM for k-shot Learning}}
    \label{alg:comptraining}
\end{algorithm*}

\section{Model Architecture}
\begin{figure}[!h]
\centering
\input{figs/transformer.tex}
\caption{Illustration of how we use transformers for sequence classification (left) and sequence transduction (right) problems.}
\label{fig:outline}
\end{figure}
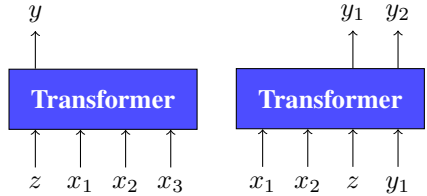

Figure \ref{fig:outline} shows an illustration of how we use transformers for sequence classification (left) and sequence transduction (right) problems.
In the classification setting the input is a sequence $(x_1\cdots x_n)$ and the output is a discrete label $y$.
In the transduction setting, the input $(x_1\cdots x_n)$ and output $(y_1\cdots y_m)$ are sequences.
$z$ is an embedding vector we refer to as the \textit{task embedding} and appears in the input to the transformer, in addition to the input sequence.
The task embedding $z$ is task specific, and is inferred on the fly for each task during training.
Learning a new task $\mathcal{T}$ at test time involves inferring the corresponding task embedding $z_\mathcal{T}$, leaving the rest of the model parameters untouched.

%% file: figs/transformer.tex
\begin{minipage}{0.21\textwidth}
\centering
\begin{tikzpicture}[
block/.style={
draw,
fill=blue,
rectangle}
]
\node (tf) [draw, minimum size=10mm, minimum width=25mm, minimum height=8mm, fill=blue!70!white, draw=black, text=white] {\textbf{Transformer}};

\coordinate[below left = 5mm and 9mm of tf.south]   (a1);
\coordinate[right = 6mm of a1]              (a2);
\coordinate[right = 6mm of a2]              (a3);
\coordinate[right = 6mm of a3]              (a4);

\foreach \i [count=\xi from 1] in {$z$, $x_1$, $x_2$, $x_3$}
\draw[->]  (a\xi) node[below] {\i} -- (a\xi|- tf.south);

\coordinate[above left = 5mm and 9mm of tf.north]   (b1);
\foreach \i [count=\xi from 1] in {$y$}
\draw[->] (tf.north -| b\xi) -- (b\xi) node[above] {\i};

\end{tikzpicture}
\end{minipage}
\begin{minipage}{0.21\textwidth}
\centering
\begin{tikzpicture}[
block/.style={
draw,
fill=blue,
rectangle}
]
\node (tf) [draw, minimum size=10mm, minimum width=25mm, minimum height=8mm, fill=blue!70!white, draw=black, text=white] {\textbf{Transformer}};

\coordinate[below left = 5mm and 9mm of tf.south]   (a1);
\coordinate[right = 6mm of a1]              (a2);
\coordinate[right = 6mm of a2]              (a3);
\coordinate[right = 6mm of a3]              (a4);

\foreach \i [count=\xi from 1] in {$x_1$, $x_2$, $z$, $y_1$}
\draw[->]  (a\xi) node[below] {\i} -- (a\xi|- tf.south);

\coordinate[above left = 5mm and -3mm of tf.north]   (b1);
\coordinate[right = 6mm of b1]  (b2);
\foreach \i [count=\xi from 1] in {$y_1$, $y_2$}
\draw[->] (tf.north -| b\xi) -- (b\xi) node[above] {\i};

\end{tikzpicture}
\end{minipage}